\documentclass{svproc}
\usepackage{url}
\usepackage{graphicx}
\usepackage{subfigure}
\usepackage{tabularx}
\usepackage{xcolor}
\newcolumntype{C}[1]{>{\centering\let\newline\\\arraybackslash\hspace{0pt}}m{#1}}

\usepackage[utf8]{inputenc}
\usepackage{hyperref}
\usepackage{amsmath,amssymb}

\begin{document}
\mainmatter              
\title{Evaluating the Reliability of CNN Models on Classifying Traffic and Road Signs using LIME}
\titlerunning{Evaluating the Reliability of CNN Models}  
%
\author{}
\author{Md. Atiqur Rahman \and Ahmed Saad Tanim \and
Sanjid Islam \and Fahim Pranto \and G. M. Shahariar \and Md. Tanvir Rouf Shawon}

\authorrunning{Md. Atiqur et al.} 
%
\tocauthor{Md. Atiqur Rahman, Ahmed Saad Tanim, Sanjid Islam, Fahim Pranto, 
G. M.Shahariar and Md. Tanvir Rouf Shawon}
\institute{Ahsanullah University of Science and Technology, Dhaka, Bangladesh\\
\email{\{ani.atikur99, ahmedsaadtanim, sanjid8426, fahimpranto002, sshibli745, shawontanvir95\}@gmail.com}}

\maketitle          

\begin{abstract}
The objective of this investigation is to evaluate and contrast the effectiveness of four state-of-the-art pre-trained models, ResNet-34, VGG-19, DenseNet-121, and Inception V3, in classifying traffic and road signs with the utilization of the GTSRB public dataset. The study focuses on evaluating the accuracy of these models' predictions as well as their ability to employ appropriate features for image categorization. To gain insights into the strengths and limitations of the model's predictions, the study employs the local interpretable model-agnostic explanations (LIME) framework. The findings of this experiment indicate that LIME is a crucial tool for improving the interpretability and dependability of machine learning models for image identification, regardless of the models achieving an f1 score of 0.99 on classifying traffic and road signs. The conclusion of this study has important ramifications for how these models are used in practice, as it is crucial to ensure that model predictions are founded on the pertinent image features.


\end{abstract}
\section{Introduction}
The continuous increment in the demand for autonomous cars is prevailing in the automobile industry in a rapid manner. The market is so vast that experts predict it will reach 3 Trillion US dollars by the year 2031\footnote{https://shorturl.at/dISU0}. This sector is entirely dependent on the application of artificial intelligence, where image segmentation and categorization are used widely. The traffic and road sign detection system is the most crucial requirement for an autonomous vehicle. The accuracy of driving in general is greatly influenced by the performance of this particular model. Several works like ~\cite{Zhang} ~\cite{wang} ~\cite{suto} have been done in this domain where authors used different variations of CNN and pre-trained CNN models. Among other technologies, computer vision plays a crucial role in the development of autonomous vehicle technology. By leveraging advanced object detection algorithms, combined with state-of-the-art ML models, autonomous vehicles can effectively recognize and interpret their surroundings, making the entire driving process safer and more efficient. One of the primary functions of computer vision in autonomous vehicles is road sign recognition. This capability allows the vehicle to accurately perceive its environment, detect potential hazards, and make informed decisions regarding its navigation. Object detection and classification are achieved using deep learning models such as Convolutional Neural Networks (CNNs), which are trained on large-scale datasets to recognize a wide range of objects with high accuracy.

CNNs (Convolutional Neural Networks) have become the most extensively used classification technique for images, such as traffic and road sign recognition, owing to their exceptional performance. However, despite their success, the interpretability of CNN models remains a critical challenge for researchers and practitioners alike. With millions of parameters, CNN models' complexity poses significant hurdles in understanding how these models function. The nonlinearity and abstract nature of CNN models further increase the difficulty in understanding their decision-making process. Moreover, CNN models are frequently trained using back-propagation, a technique that makes it impossible for humans to track the millions of parameters involved. Consequently, the lack of a straightforward mathematical formula for the logic behind these models' judgments often results in a failure to comprehend the decision-making process of the model, limiting its dependability and trustworthiness. As a result, researchers have turned their attention to developing new methods to interpret CNN models, such as visualizing activation maps and feature visualization techniques. These techniques help researchers to understand the underlying decision-making process of CNN models, providing more insight into their functionality and ultimately increasing their dependability and trustworthiness.
Local Interpretable Model-Agnostic Explanations (LIME~\cite{rib:sin}) is a prominent approach for analyzing machine learning model predictions, such as CNNs. By approximating the model's behavior in the area of the prediction, LIME produces local explanations for individual forecasts. LIME can thus give insights into how the CNN model produces classification judgments.

This study focuses on the evaluation of the Convolutional Neural Network (CNN) models' reliability in the identification of traffic and road signs. To assess the dependability of these models, we employ Local Interpretable Model-Agnostic Explanations (LIME) and investigate how effectively LIME can elucidate CNN model decisions. Our research endeavors to determine whether these explanations provide valuable insights into the CNN models' behavior, aiding researchers in understanding how these models function and ultimately improving their dependability. By utilizing LIME, we seek to evaluate the interpretability of CNN models in a thorough and comprehensive manner, allowing us to draw meaningful conclusions about their performance and applicability in real-world scenarios 

To do this, we used pre-trained models~\cite{he:zha,sim:zis,sze:van}, namely VGG-19, ResNet-34, DenseNet-121, and Inception V3, which were trained on a vast dataset of traffic and road signs which is German Traffic Sign Recognition Benchmark~\cite{hoben:stal}. We utilize LIME to construct explanations for individual predictions generated by these models and assess their quality using a variety of measures.

This work provides the following contributions. 
\begin{itemize}
    \item Utilize the GTSRB dataset to classify images
    \item Classify the images using four pre-trained models: ResNet-34, VGG-19, Inception V3, and DenseNet-121
    \item Evaluate the results of the pre-trained models
    \item Give a thorough examination of the reliability of those CNN pre-trained models in identifying traffic and road signs using LIME.
    \item The research will assist in increasing the reliability and trustworthiness of CNN models for traffic and road sign categorization, hence improving the safety of autonomous cars and intelligent transportation systems.
\end{itemize}

The study is structured as follows: 
Section \ref{section2} encompasses a detailed review of the relevant literature. In Section \ref{section3}, the background study is extensively elaborated, providing an in-depth understanding of the context. Section \ref{section4} offers a comprehensive description of the dataset used in the research. The proposed approach is discussed in Section \ref{section5}, which outlines the methodology and its rationale in detail. In Section \ref{section6}, the experiment's results are presented and analyzed, providing valuable insights into this research outcome. Section \ref{section7} briefs the interpretation of models' output using LIME. Finally, in Section \ref{section8}, the study concludes, summarizing the main findings, highlighting their significance, and pointing to future directions for research in the field.

\section{Related Works}
\label{section2}
Over the past few years, several studies have been done on detecting traffic and road signs. We are primarily concentrating on works that used LIME to analyze deep learning models along with pre-trained CNN models. Given the circumstances, this section is separated into two subsections that contain papers on pre-trained models and papers on LIME.

\subsection{Pretrained models for road sign detection} 
Jingwei et al.~\cite{cao:son} constructed a traffic sign identification system that obtained $99.75\%$ accuracy while Gu and Si~\cite{gu:si} suggested a lightweight integration framework based on YOLOv4 for real-time traffic sign identification, attaining a mean average precision (mAP) score of $80.62\%$ on the GTSDB~\cite{hoben:stal} dataset. Soulef et al.~\cite{bou:mess} employed VGG-16, VGG-19, AlexNet, and Resnet-50 to recognize traffic and road signs, with AlexNet achieving the greatest average accuracy, recall, and f1-score values on the GTSRB dataset of $96.39\%$, $94.88\%$, and $95.3\%$ respectively. Houben et al.~\cite{hoben:stal} conducted an objective evaluation of different algorithms to identify traffic signs in images captured from the real world, achieving detection rates of $91.3\%$ for prohibit signs, $90.7\%$ for danger signals, and $69.2\%$ for mandatory signs using the HOG with LDA method. Hough-like detection rates for prohibit, danger, and mandatory signs were $55.3\%$, $65.1\%$, and $34.7\%$ respectively. An Indian Cautionary Traffic Sign (ICTS) data set was created by Satish et al.~\cite{satti:sk}. that contains 19,775 images that were obtained through difficult circumstances. The dataset has been classified into 40 groups. To classify those images, they used Vgg-16, LeNET, ResNet-50 and AlexNet. They achieved the highest accuracy of 98.73\% using ResNet-50. Soufiane et al.~\cite{naim:mou} presented an extremely light deep cnn architecture which they call LiteNet to classify the road signs for traffic in the GTSRB dataset. Using this network they achieved an accuracy of 99.15\% that outperformed many other architectures. A novel, VGG-based network architecture with fewer parameters is proposed by AttoumaneIt et al.~\cite{lok:gra} The network architecture comprises an input layer, followed by six convolutional layers, three layers that incorporate maximum pooling, and a single layer that is fully connected with batch normalization and dropout, and one softmax layer as the output. Using the proposed model, they were able to achieve an accuracy of 99.33\%.

\subsection{Usage of Lime in image classification}
LIME is used in many papers to explain the model predictions. Schneider et al~\cite{sch:mul} used LIME to justify if the feature of the images is marked as relevant. Sahay et al.~\cite{sah:oma} explained the captions generated by the model using LIME. Magesh et al.~\cite{mag:myl} utilizes LIME explanations to distinguish PD from non-PD, using visual superpixels on the DaTSCANs. Chen et al.~\cite{chn:lee} utilized LIME to show relevant  ROI technique, and this was also produced with U-Net segmentation, which deleted non-lung regions of the pictures to prevent the classifier from being sidetracked by insignificant characteristics. Rajaraman et al.~\cite{raj:can} underlined the advantages of showing and explaining CNN activations and predictions in the context of the pediatric chest radiograph pneumonia detection challenge.


\section{Background Study}
\label{section3}
This section aims to present a concise overview of the pre-trained models, lime image explainer, and some performance evaluation metrics utilized in our experiments.

\subsection{Pre-trained Models}

\paragraph{}
\textbf{ResNet:} The ResNet deep CNN architecture was first described in Deep Residual Learning for Image Recognition by He et al.~\cite{he:zha}. It is a variation of the ResNet design, known for its capacity to train very deep networks while not encountering the issue of vanishing gradients. The general architecture of ResNet (Residual Network) contains Residual Blocks and has the added functionality of `skip connections', which are used to skip certain layers to counter this issue. This allows the performance of the ResNet architecture to not degrade as more layers are added on top of it, so compared to plain neural networks. It has been demonstrated that ResNet performs extremely well when used in image recognition tasks, as it has achieved great performance results on benchmarks of ImageNet, CIFAR-10, and CIFAR-100, and as such it is very popular in computer vision tasks. Compared to VGG, it performs faster because of its simpler architecture.

\textbf{VGG:} VGG~\cite{sim:zis} is a CNN that is taught using the ImageNet dataset~\cite{den:don}, a variation of the VGG network architecture that has numerous completely connected layers after a series of convolution and max pooling layers. At the University of Oxford, the Visual Geometry Group first presented the architecture in 2014, where it outperformed other cutting-edge models and is still utilized in object detection, semantic segmentation and also image production, etc.

\textbf{DenseNet} DenseNet is a CNN architecture that was introduced in 2016 by Gao Huang et al.~\cite{zhu:new} from the Chinese University of Hong Kong. It is renowned for having a dense connectivity pattern in which all of the layers are connected to one another in a feed-forward manner. This contrasts with traditional CNNs, where each layer is only connected to a subset of the previous layers. DenseNet's design is made up of a number of dense blocks, each of which has numerous levels. Each layer takes as inputs the feature maps of all layers before it, and then it concatenates those feature maps with its own feature maps. 

\textbf{Inception} Inception is a convolutional neural network architecture, introduced by Szegedy et al.~\cite{sze:liu}, for image classification. It was trained on the ImageNet dataset and achieved a cutting-edge performance on the image classification task. It is designed to improve the efficiency of the network by reducing the number of parameters required, while still maintaining high accuracy. The Inception architecture achieves this by using multiple layers of filters with different sizes to capture features at different scales.

\subsection{LIME}

In order to comprehend how any black box (abstract) ML model predicts, researchers initially developed LIME~\cite{rib:sin}(Local Interpretable Model-agnostic Explanations), during the application of local surrogate models. LIME is a method where it explains each individual prediction of any black box model by a locally interpretable (surrogate) model. To make this work, from modified samples and the corresponding black box model predictions, LIME generates a new sample. This sample dataset trains an easy-to-interpret model that delivers accurate local approximations of the predictions from the ML model and is weighted by the fresh sampled examples based on their closeness to the point of interest.
Equation~\ref{lime_eqn} is the mathematical formula of Lime.
\begin{equation}
\label{lime_eqn}
    \xi(x) = \mathop{\arg \min}\limits_{g \in G} L(f, g, \pi_{x}) + \Omega(g)
\end{equation}
Here,
$\xi(x)$ is known as the approximate model that LIME generates for a particular instance $x$.
$\xi$ is the black-box model being explained.
$G$ is the interpretable model in the set that LIME can use to approximate $\xi$. $L(f, g, \pi_{x})$ is the loss function that calculates the disparity between the model's predictions and the actual values of the black-box model $\xi$ and the interpretable model $g$ on the instance $x$, weighted by the proximity of $x$ to the instances in the dataset. $\pi_{x}$ is the proximity kernel that determines the weighting of the instances in the dataset depending on their resemblance with the example $x$.
$\Omega(g)$ is the complexity penalty that encourages the interpretable model $g$ to be simple and interpretable.

\subsection{Performance Metrics}
\label{sec:performance_metrics}
In this experiment, we evaluated the models' performance using various performance evaluation metrics such as accuracy, recall, f1-score, precision, roc-auc score, and Mathews correlation coefficient. Accuracy is commonly used to assess a classifier's classification performance by researchers. Precision measures the proportion of correct positive predictions among all predicted positives and is particularly useful when dealing with imbalanced class distributions. The percentage of correct positive predictions among all real positives is calculated by the recall. The harmonic mean of accuracy and recall is known as F1-score, indicating the models' overall effectiveness. ROC AUC value represents the models' ability to distinguish between positive and negative classifications, with higher AUC indicating better performance. An AUC score of 1 means perfect classification accuracy. The Matthews correlation coefficient is a balanced metric used to evaluate binary and multi-class classifications that consider true and false positives and negatives, making it suitable even for imbalanced class sizes.


\section{Dataset Description}
\label{section4}
\begin{figure}[h]
    \centering
    \begin{subfigure}
    \centering
    \includegraphics[width=2.3cm, height=2.3cm]{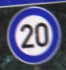}
    \end{subfigure}
    \begin{subfigure}
    \centering
    \includegraphics[width=2.3cm, height=2.3cm]{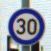}
    \end{subfigure}
    \begin{subfigure}
    \centering
    \includegraphics[width=2.3cm, height=2.3cm]{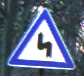}
    \end{subfigure}
    \begin{subfigure}
    \centering
    \includegraphics[width=2.3cm, height=2.3cm]{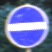}
    \end{subfigure}
    \begin{subfigure}
    \centering
    \includegraphics[width=2.3cm, height=2.3cm]{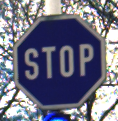}
    \end{subfigure}
    \begin{subfigure}
    \centering
    \includegraphics[width=2.3cm, height=2.3cm]{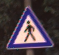}
    \end{subfigure}
    \begin{subfigure}
    \centering
    \includegraphics[width=2.3cm, height=2.3cm]{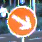}
    \end{subfigure}
    \begin{subfigure}
    \centering
    \includegraphics[width=2.3cm, height=2.3cm]{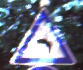}
    \end{subfigure}
    \caption{Samples from the dataset}
    \label{fig:sample}
\end{figure}
For our experiment, we have also used GTSRB ~\cite{hoben:stal} dataset. The dataset contains a total of $51,839$ images, all of which are in RGB format. Out of these, $39,209$ images are used for training and  $12,630$ images are utilized for generating predictions. The images are categorized into $43$ distinct classes, and their dimensions range from $25 \times 25$ to $266 \times 232$. The image depicted in Figure~\ref{fig:sample} provides a preview of some samples included in this dataset. It should be noted that the dataset contains numerous versions of the same signs, which varies in terms of lighting, focus, position, and other factors. 
\begin{figure}[h]
    \centering
    \includegraphics[width=\textwidth, height=6cm]{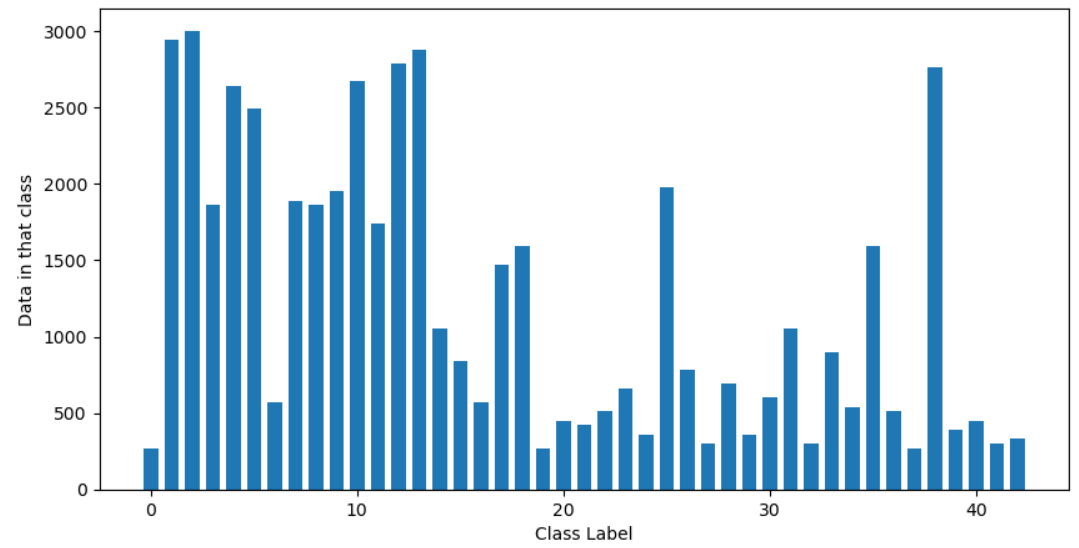}
    \caption{Class distribution}
    \label{fig:class_distribution}
\end{figure}  
The distribution of data in the dataset is not balanced, with some classes having more than $2500$ images while others having less than $500$ images. Figure~\ref{fig:class_distribution} depicts the distribution of classes.

\section{Proposed Approach}
\label{section5}
\begin{figure}[h]
  \centering
  \includegraphics[width=8cm, height=11cm]{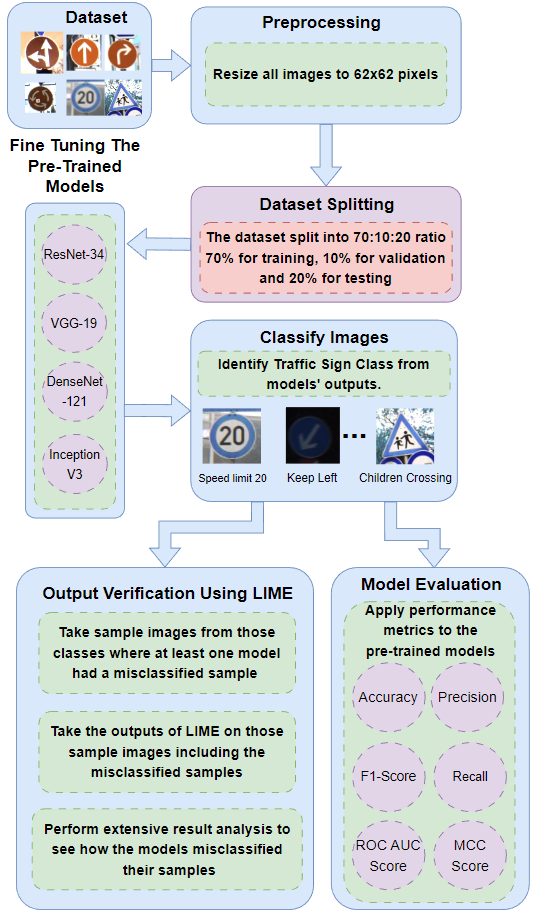}
  \caption{Proposed Methodology}
  \label{proposed_methodology}
\end{figure}

Figure~\ref{proposed_methodology} presents the proposed methodology of this experiment. The methodology is divided into five phases. The phases are described below:

\textbf{Phase I:} This phase included dataset preparation. The images of the dataset are not similar in size. We resized the images to $62 \times 62$ pixels. Initially, the dataset is split into two parts as follows: train sets and test sets. There were no validation data. We concatenated both the train sets and test sets and separated the concatenated datasets into train, validation, and test with a ratio of $70:10:20$ respectively with stratification to ensure that all the class levels are distributed properly.

\textbf{Phase II: } This phase involves training all the models utilized in the current study. To accomplish this, we utilize the training set to train four pre-trained CNN models, ResNet-34, VGG-19, DenseNet-121, and Inception V3. To monitor the models' performance and avoid overfitting, we calculate the validation loss and accuracy after each epoch using the validation dataset. This rigorous evaluation approach ensures that the models are thoroughly evaluated and that their performance is well-documented throughout the training process.

\textbf{Phase III: } On the test dataset, we calculate the models' classification performance. We assess the effectiveness of the models using the performance metrics outlined in subsection~\ref{sec:performance_metrics}. We used traditional performance metrics which are used for classification tasks. 

\textbf{Phase IV: } In this phase, we employed LIME to generate visual explanations for a small number of images from the test set, which help to explain the rationale behind the models' predictions.

\section{Experimental Results}
\label{section6}
The result of the experiments that were conducted are presented in this section. Our proposed approach utilized four pre-trained models, so we performed four separate experiments.

\subsection{Hyper-parameters Setting}
We have experimented with different hyper-parameters to achieve the best performance of each pre-trained model. We have fine-tuned all the models for $10$ epochs, AdamW optimizer with $0.0001$ learning rate, and of $8$ batch size. Additionally, we applied random rotation and resize the images. For ResNet-34, VGG-19, and DenseNet-121 the images are resized to $224 \times 224$, and in the case of Inception-V3, the images are resized to a dimension of $299 \times 299$. The random rotation and resize are performed by using PyTorch Torchvision library. The number of resulting images is unaffected by the augmentation because only one augmentation is done per image into the custom dataloader on the fly. We have used the PyTorch framework for training and testing the models.

\subsection{Classification Performance}
According to the experimental results, all four pre-trained models in the proposed methodology exhibit excellent performance across all evaluation metrics, with scores consistently near $1$. This suggests that the models can perform accurate classification of the data with high consistency and low levels of false positives and negatives. A high F1 score implies a good mix of precision and recall, meaning that the models can detect the most relevant cases while minimizing false alarms. Overall, these results suggest that the models are dependable and efficient for their intended task. The macro average scores can be found in Table~\ref{table:results}.

\begin{table}[h]
\centering
\caption{The evaluation metrics of the pre-trained models' performance}
\fontsize{8.5pt}{11pt}
\selectfont
\begin{tabular}{|C{2.2cm}|C{1.8cm}|C{1.7cm}|C{1.3cm}|C{1.3cm}|C{1.3cm}|C{1.3cm}|}
\hline
\textbf{\begin{tabular}[c]{@{}c@{}}Pre-trained\\ model\end{tabular}} & \textbf{Accuracy (\%)} & \textbf{Precision} & \textbf{Recall} & \textbf{F1-Score} 
& \textbf{MCC-Score} & \textbf{ROC-AUC Score} \\ \hline
\textbf{DenseNet-121}  & 99.95 & 0.9990 & 0.9995 & 0.9992 & 0.9989 & 0.9999\\ \hline
\textbf{Inception V3} & 99.92          & 0.9982          & 0.9992          & 0.9987 & 0.9987 & 0.9999\\ \hline
\textbf{ResNet-34}    & 99.94          & 0.9992          & 0.9994          & 0.9993   & 0.9991 & 0.9999\\ \hline
\textbf{VGG-19}       & 99.66          & 0.9971          & 0.9966          & 0.9968  & 0.9971 & 0.9999\\ \hline
\end{tabular}
\label{table:results}
\end{table}

The experimental results show that each pre-trained model in the proposed methodology has a high level of performance across all evaluation metrics. The metrics are all above 99\%. This indicates that the model can correctly classify the data with a high level of consistency and that there is a low level of false positives and false negatives. The high F1 score in particular suggests that the balance between precision and recall is favorable, meaning the model is able to identify the majority of relevant cases while minimizing false alarms. Overall, the results show that the models are reliable and effective for the task it was designed for.

\begin{figure}[h]
    \centering
    \setcounter{subfigure}{0}
    \subfigure[DenseNet-121]
    {
        \includegraphics[width=0.47
\textwidth]{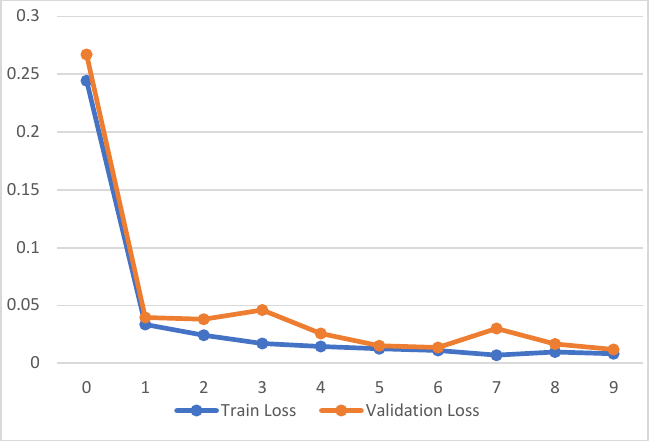}
        \label{fig:densenet_loss}
    }
    \subfigure[Inception V3]
    {
        \includegraphics[width=0.47\textwidth]{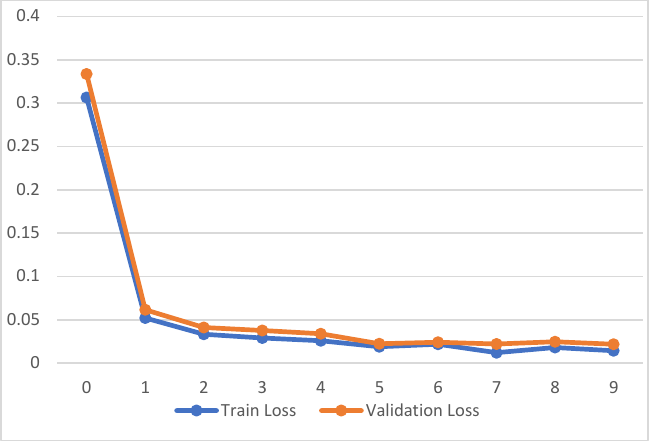}
        \label{fig:inception_loss}
    }
    \subfigure[ResNet-34]
    {
        \includegraphics[width=0.47\textwidth]{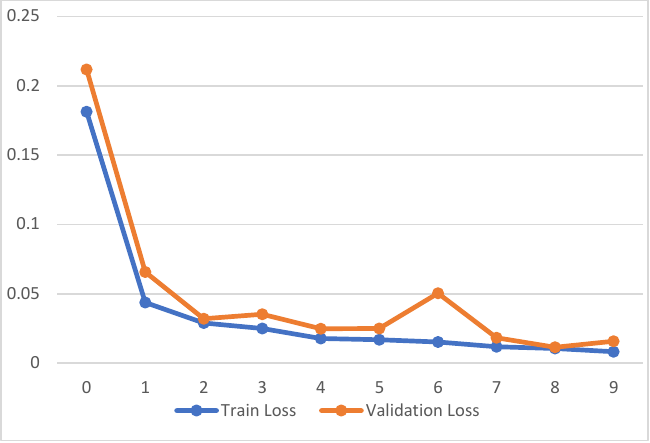}
        \label{fig:resnet_loss}
    }
    \subfigure[VGG-19]
    {
        \includegraphics[width=0.47\textwidth]{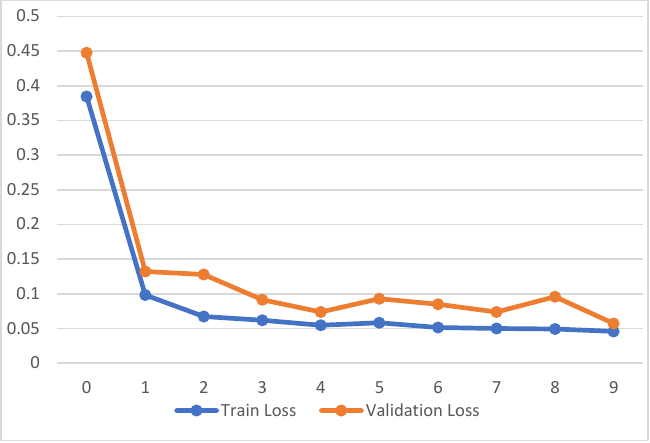}
        \label{fig:vgg_loss}
    }
    \caption{Training and validation loss 
 of four models.~\ref{fig:densenet_loss} to  \ref{fig:vgg_loss} are loss curve of DenseNet-121, Inception V3, ResNet-34 and VGG-19 respectively.}
\label{loss_curves}
\end{figure}

The figure~\ref{loss_curves} represents the training and validation loss curve of each of the models. From the loss curve, it is observed that the training and validation loss decreases, indicating that none of the models are overfitted. There are significant decreases in training loss after the first epoch for all four models. A small fluctuation is noticed in the validation loss curve of the ResNet-34 model after at 6th epoch.

\section{Model Interpretation using LIME}
\label{section7}
The colorful regions in the images demonstrate the features the models used to classify these images. The green regions indicate that the model considered these regions to be heavily responsible for the positive classification of the image. In contrast, the red regions indicate that this region accounted for the negative classification of the image. Among the 43 classes, we used sample images that have been misclassified by at least one model. Although LIME is said to be not stable, it is consistent in giving similar answers. This is verified by re-running LIME several times on the sample images. The detailed analysis using LIME of the misclassified images is presented below.

\begin{figure}[!htbp]
    \centering
    \setcounter{subfigure}{0}
    \subfigure[Inception V3]
    {
        \includegraphics[width=2.5cm, height=2.5cm]{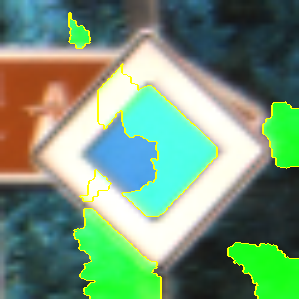}
        \label{prio1}
    }
    \subfigure[DenseNet-121]
    {
        \includegraphics[width=2.5cm, height=2.5cm]{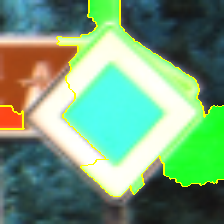}
        \label{prio2}
    }
    \subfigure[VGG-19]
    {
        \includegraphics[width=2.5cm, height=2.5cm]{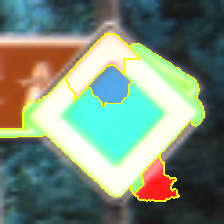}
        \label{prio3}
    }
    \subfigure[ResNet-34]
    {
        \includegraphics[width=2.5cm, height=2.5cm]{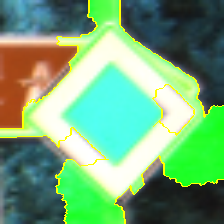}
        \label{prio4}
    }
    \subfigure[Inception V3]
    {
        \includegraphics[width=2.5cm, height=2.5cm]{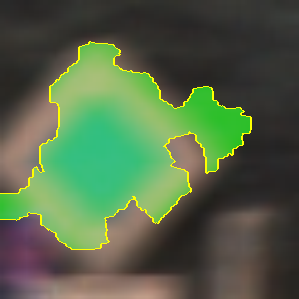}
        \label{prio5}
    }
    \subfigure[DenseNet-121]
    {
        \includegraphics[width=2.5cm, height=2.5cm]{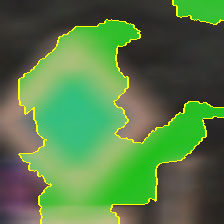}
        \label{prio6}
    }
    \subfigure[VGG-19]
    {
        \includegraphics[width=2.5cm, height=2.5cm]{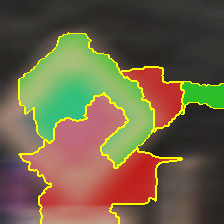}
        \label{prio7}
    }
    \color{black}
    \subfigure[ResNet-34]
    {
        \includegraphics[width=2.5cm, height=2.5cm]{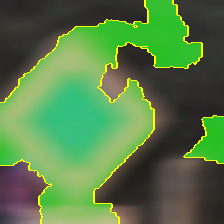}
        \label{prio8}
    }
    \color{black}
    \caption{LIME interpretation of models' output for Priority Road class where \ref{prio1}, \ref{prio2}, \ref{prio3}, \ref{prio4}, \ref{prio5}, \ref{prio6} and \ref{prio8} are correctly classified and \ref{prio7} misclassified}
    \label{priority_road}
\end{figure}
Here,~\ref {prio1} to~\ref{prio4} images in Figure~\ref{priority_road} are correctly classified samples for the class `Priority Road'. By observing the LIME markings generated on them we can see that in DenseNet-121 and ResNet-34, the superpixel regions are covering the traffic signs perfectly. In Inception V3 and VGG-19, the superpixel regions are marking them partially. In the image~\ref{prio7} we can see that VGG-19 misclassified a sample from the same class due to its inability to identify the regions correctly. Comparing images~\ref{prio3} and~\ref{prio7} we can see that, with reason, VGG-19 could not focus the same regions for the same class as the image~\ref{prio7} is blurry compared to image~\ref{prio3}. The rest of the models almost perfectly covered the correct regions thus their correct classifications were justified. 

\begin{figure}[!htbp]
    \centering
    \setcounter{subfigure}{0}
    \subfigure[Inception V3]
    {
        \includegraphics[width=2.5cm, height=2.5cm]{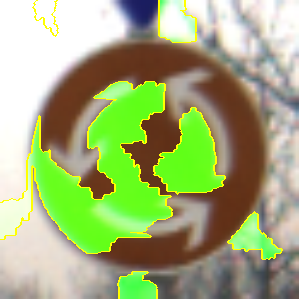}
        \label{round1}
    }
    \subfigure[DenseNet-121]
    {
        \includegraphics[width=2.5cm, height=2.5cm]{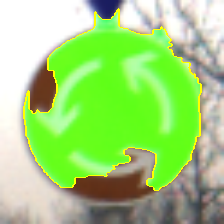}
        \label{round2}
    }
    \subfigure[VGG-19]
    {
        \includegraphics[width=2.5cm, height=2.5cm]{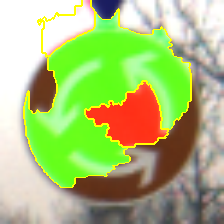}
        \label{round3}
    }
    \subfigure[ResNet-34]
    {
        \includegraphics[width=2.5cm, height=2.5cm]{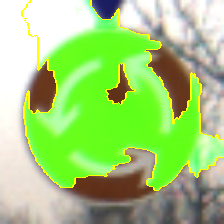}
        \label{round4}
    }
    \subfigure[Inception V3]
    {
        \includegraphics[width=2.5cm, height=2.5cm]{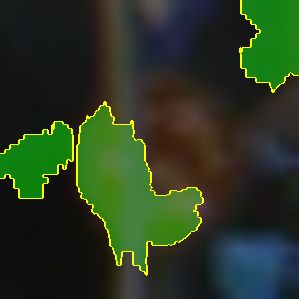}
        \label{round5}
    }
    \color{black}
    \subfigure[DenseNet-121]
    {
        \includegraphics[width=2.5cm, height=2.5cm]{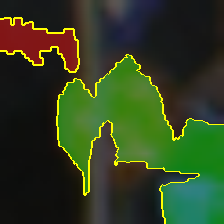}
        \label{round6}
    }
    \subfigure[VGG-19]
    {
        \includegraphics[width=2.5cm, height=2.5cm]{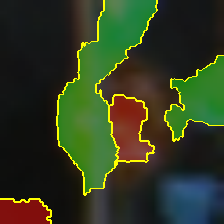}
        \label{round7}
    }
    \subfigure[ResNet-34]
    {
        \includegraphics[width=2.5cm, height=2.5cm]{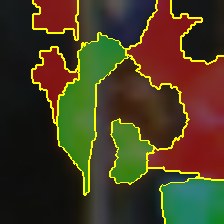}
        \label{round8}
    }
    \color{black}
    \caption{LIME interpretation of models' output for Roundabout Mandatory class where \ref{round1}, \ref{round2}, \ref{round3}, \ref{round4} and \ref{round6} are correctly classified and \ref{round5}, \ref{round7}, and  \ref{round8} are misclassified}
    
\end{figure}
These samples are for the class `Roundabout Mandatory'. Here the images~\ref{round1},~\ref{round4} and ~\ref{round2},~\ref{round3} are the correctly classified samples from each of the models. But the superpixel regions for the respective models are not near as good as in the misclassified images~\ref{round5},~\ref{round7} and~\ref{round8}. Here in image~\ref{round6}, DenseNet-121 could classify correctly, as it classified the image using the right regions, so this classification is reliable.

\begin{figure}[!htbp]
    \centering
    \setcounter{subfigure}{0}
    \subfigure[Inception V3]
    {
        \includegraphics[width=2.3cm, height=2.1cm]{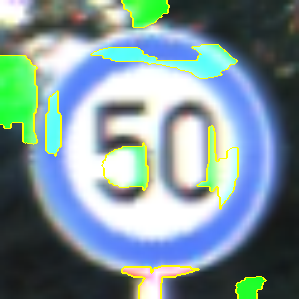}
        \label{speed_20_1}
    }
    \subfigure[DenseNet-121]
    {
        \includegraphics[width=2.3cm, height=2.1cm]{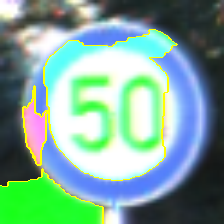}
        \label{speed_20_2}
    }
    \subfigure[VGG-19]
    {
        \includegraphics[width=2.3cm, height=2.1cm]{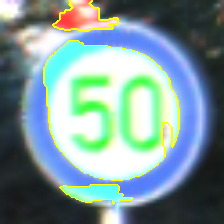}
        \label{speed_20_3}
    }
    \subfigure[ResNet-34]
    {
        \includegraphics[width=2.3cm, height=2.1cm]{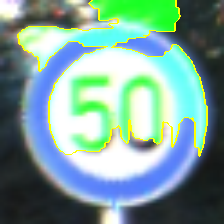}
        \label{speed_20_4}
    }
    \subfigure[Inception V3]
    {
        \includegraphics[width=2.3cm, height=2.1cm]{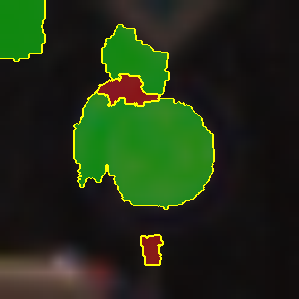}
        \label{speed_20_5}
    }
    \subfigure[DenseNet-121]
    {
        \includegraphics[width=2.3cm, height=2.1cm]{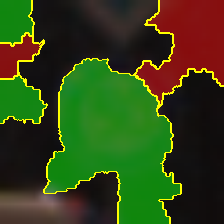}
        \label{speed_20_6}
    }
    \color{black}
    \subfigure[VGG-19]
    {
        \includegraphics[width=2.3cm, height=2.1cm]{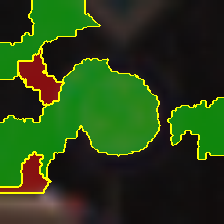}
        \label{speed_20_7}
    }
    \subfigure[ResNet-34]
    {
        \includegraphics[width=2.3cm, height=2.1cm]{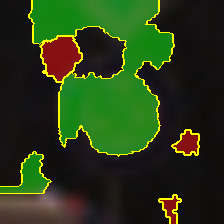}
        \label{speed_20_8}
    }
    \color{black}
    \caption{LIME interpretation of models' output for the Speed limit (50km/h) class where \ref{speed_20_5}, \ref{speed_20_7}, \ref{speed_20_8} and \ref{speed_20_4}, \ref{speed_20_3}, \ref{speed_20_2}, \ref{speed_20_1} are correctly classified and \ref{speed_20_6} misclassified}
    \label{fig:speed_limit_50}
\end{figure}

From the class `Speed Limit 50 kmph' shown in Fig~\ref{fig:speed_limit_50}, the sample images~\ref{speed_20_1},~\ref{speed_20_4} and ~\ref{speed_20_2}, ~\ref{speed_20_3} highlighted the superpixel regions mostly inside the road sign resulting in clear positive predictions. On the contrary, in images~\ref{speed_20_5}, ~\ref{speed_20_7}, ~\ref{speed_20_8} and ~\ref{speed_20_6} the positive superpixel regions are scattered along with the presence of negative regions. Due to its unclear and dark pixels and the sign being not visible, DenseNet-121 failed to correctly classify that sample image and the rest of the models classified this image using the wrong superpixel regions. 

\begin{figure}[h]
    \centering
    \setcounter{subfigure}{0}
    \subfigure[Inception V3]
    {
        \includegraphics[width=2.3cm, height=2.1cm]{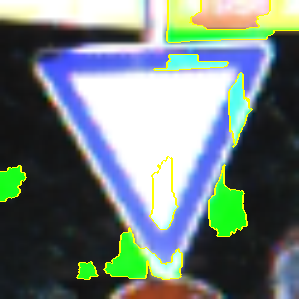}
        \label{yield1}
    }
    \subfigure[DenseNet-121]
    {
        \includegraphics[width=2.3cm, height=2.1cm]{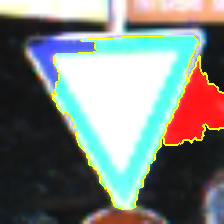}
        \label{yield2}
    }
    \subfigure[VGG-19]
    {
        \includegraphics[width=2.3cm, height=2.1cm]{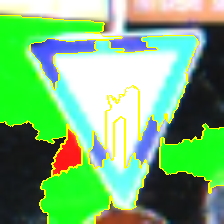}
        \label{yield3}
    }
    \subfigure[ResNet-34]
    {
        \includegraphics[width=2.3cm, height=2.1cm]{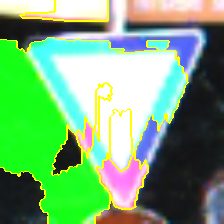}
        \label{yield4}
    }
    \subfigure[Inception V3]
    {
        \includegraphics[width=2.3cm, height=2.1cm]{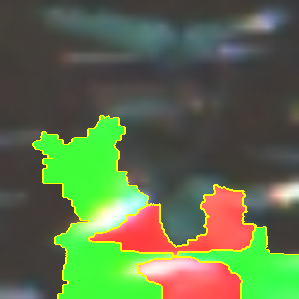}
        \label{yield5}
    }
    \subfigure[DenseNet-121]
    {
        \includegraphics[width=2.3cm, height=2.1cm]{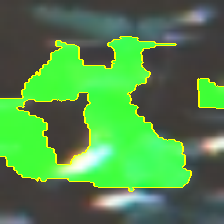}
        \label{yield6}
    }
    \color{black}
    \subfigure[VGG-19]
    {
        \includegraphics[width=2.3cm, height=2.1cm]{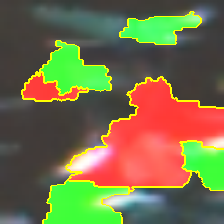}
        \label{yield7}
    }
    \subfigure[ResNet-34]
    {
        \includegraphics[width=2.3cm, height=2.1cm]{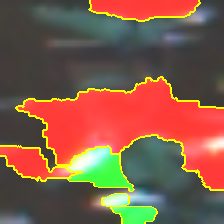}
        \label{yield8}
    }
    \color{black}
    \caption{LIME interpretation of models' output for Yield class where \ref{yield7} and \ref{yield1}, \ref{yield4} and \ref{yield2}, \ref{yield3} are correctly classified and \ref{yield8}, \ref{yield6} and \ref{yield5} are misclassified}
\end{figure}
Lastly, the class `Yield' provides a very interesting LIME explanation where it is clear that even for the correctly classified cases, the models' Inception V3, VGG-19, and ResNet-34 used the wrong image features along with some correct regions to classify. This strategy backfired when the image is blurry, dark, and very unclear as visualized in the images~\ref{yield5}, ~\ref{yield7}, ~\ref{yield8} and ~\ref{yield6}. 

This clarifies that VGG-19 is the only one to correctly classify this image but using all the wrong regions. DenseNet-121 was more reliable in this case as it marked the sign comparatively better but still failed to classify as the image itself is almost beyond recognition. From studying all of the samples and which pixels the models used for each sample, we can imply that DenseNet-121 performs fairly well for all cases. However, ResNet-34 and VGG-19 classifying images at extreme conditions seem to lack compared to the other two models, despite both of them having very high-performance scores. Inception V3 mostly classifies using wrong regions and lastly, ResNet-34 can be a good alternative for DenseNet-121 in general cases as both of the models use similar image features to classify.

\section{Conclusion And Future Works}
\label{section8}

In conclusion, our study has demonstrated the potential of using the LIME framework to assess the performance of CNN classification models on the German GTSRB public dataset. This is done by firstly training the models to recognize the German GTSRB dataset, and then secondly using LIME to get a visual representation of how each model is making their prediction based on important feature regions of the images. Lastly, we take a few sample images, and their misclassified counterparts (if any) and extensively analyze and compare the different important image feature regions used by the different models for their classification. This gives us a general understanding of the real-world reliability of these models, thus making our findings have significant ramifications for the deployment of these models in real-world situations, where it is crucial to ensure that predictions are based on accurate features of the images. There are several ambitions that we would like to pursue based on this work, such as extending to cover additional image classification models and datasets, optimizing and refining the use of LIME to provide even more insightful explanations of the models' predictions, expanding the study to consider the effects of different image pre-processing techniques, and exploring and evaluating practical applications of these models in real-world scenarios for assessing their effectiveness in solving real-world problems (i.e. self-driving cars). We also have the ambition to propose a new model in the future that not only classifies images accurately but also classifies them with a high chance of relying on the correct features of the images. These future works will help to advance our understanding of CNN models and improve their performance in practical applications.

 
%
%

\end{document}